%% file: ijcai26.tex
\documentclass{article}
\usepackage{ijcai26}

\usepackage{times}
\usepackage{soul}
\usepackage{url}
\usepackage[hidelinks]{hyperref}
\usepackage[utf8]{inputenc}
\usepackage[small]{caption}
\usepackage{graphicx}
\usepackage{amsmath}
\usepackage{amsthm}
\usepackage{booktabs}
\usepackage{algorithm}
\usepackage{algorithmic}
\usepackage[switch]{lineno}
\usepackage{multirow}
\usepackage{makecell}  
\usepackage{graphicx}  
\usepackage{pifont}    
\newcommand{\cmark}{\ding{51}} 
\newcommand{\xmark}{\ding{55}} 
\usepackage{tabularx}
\usepackage{cuted}
\usepackage{caption}

\usepackage{xcolor}
\usepackage{fontawesome5}
\usepackage{hyperref}

\definecolor{linkblue}{RGB}{0, 0, 238}

\hypersetup{
    colorlinks=true,
    urlcolor=linkblue,
    linkcolor=linkblue,
    citecolor=linkblue
}

\usepackage{booktabs}
\usepackage{array}
\usepackage{ragged2e}
\usepackage[table]{xcolor} 
\usepackage{booktabs}     
\usepackage{pifont}        
\usepackage{hyperref}
\usepackage{fontawesome5} 

\usepackage{booktabs}
\usepackage{array}
\usepackage{ragged2e}
\usepackage{xcolor}
\usepackage[numbers]{natbib} 
\definecolor{rowblue}{RGB}{236, 242, 255}
\definecolor{rowgray}{gray}{0.95}
\definecolor{sheader}{RGB}{248, 242, 255}

\title{From Human Videos to Robot Manipulation: A Survey on Scalable Vision-Language-Action Learning with Human-Centric Data}

\author{
Zhiyuan Feng$^{1,5*}$
\ \ Qixiu Li$^{1,5*}$
\ \ Huizhi Liang$^{1,5*}$
\ \ Rushuai Yang$^{2,5*}$
\ \ Yichao Shen$^{3,5*}$ \\
\ \ Zhiying Du$^{4,5*}$
\ \ Zhaowei Zhang$^{5,6*}$ 
\ \ Yu Deng$^{5\dagger}$
\ \ Li Zhao$^{5}$
\ \ Hao Zhao$^{1}$
\ \ Zongqing Lu$^{6}$ \\
\ \ Oier Mees$^{7}$ 
\ \ Marc Pollefeys$^{7}$
\ \ Jiaolong Yang$^{5\dagger}$
\ \ Baining Guo$^{5}$ 
\affiliations
$^{1}$Tsinghua University \quad
$^{2}$HKUST \quad
$^{3}$Xi'an Jiaotong University \quad
$^{4}$Fudan University \\
$^{5}$Microsoft Research Asia \quad
$^{6}$Peking University \quad
$^{7}$Microsoft Zurich \\[0.35em]
\href{https://aaronfengzy.github.io/HumanCentricToVLA-Survey/}
{\faHome\ Project Page}
\quad\quad
\href{https://github.com/AaronFengZY/HumanCentricToVLA-Survey}
{\faGithub\ Repo}
}

\begin{document}

\maketitle

\begingroup
\renewcommand{\thefootnote}{\fnsymbol{footnote}}
\footnotetext[1]{Work done during internship at Microsoft Research Asia.}
\footnotetext[2]{Corresponding authors.}
\endgroup
\setcounter{footnote}{0}

\input{sections/00_abstract}
\input{sections/01_introduction}
\input{sections/02_background}
\input{sections/03_taxonomy}
\input{sections/04_datasets}
\input{sections/05_challenges}
\input{sections/06_conclusion}

\bibliographystyle{named}
\begingroup
\bibliography{ijcai26}
\endgroup

\end{document}

%% file: sections/00_abstract.tex
\begin{abstract}
    Recent progress in generalizable embodied control has been driven by large-scale pretraining of Vision–Language–Action (VLA) models. However, most existing approaches rely on large collections of robot demonstrations, which are costly to obtain and tightly coupled to specific embodiments. Human videos, by contrast, are abundant and capture rich interactions, providing diverse semantic and physical cues for real-world manipulation. Yet, embodiment differences and the frequent absence of task-aligned annotations make their direct use in VLA models challenging. This survey provides a unified view of how human videos are transformed into effective knowledge for VLA models. We categorize existing approaches into four classes based on the action-related information they derive: (i) latent action representations that encode inter-frame changes; (ii) predictive world models that forecast future frames; (iii) explicit 2D supervision that extracts image-plane cues; and (iv) explicit 3D reconstruction that recovers geometry or motion. Beyond this taxonomy, we highlight three key open challenges in this area: structuring unstructured videos into training-ready episodes, grounding video-derived supervision into robot-executable actions under embodiment and viewpoint heterogeneity, and designing evaluation protocols that better predict real-world deployment performance and transfer efficiency, thereby informing future research directions.  A curated list of papers and resources is available at \url{https://github.com/AaronFengZY/HumanCentricToVLA-Survey}.
\end{abstract}

%% file: sections/01_introduction.tex
\section{Introduction}

\begin{figure}[t!]
    \centering
    \captionsetup{type=figure, hypcap=false}
    \includegraphics[width=\columnwidth]{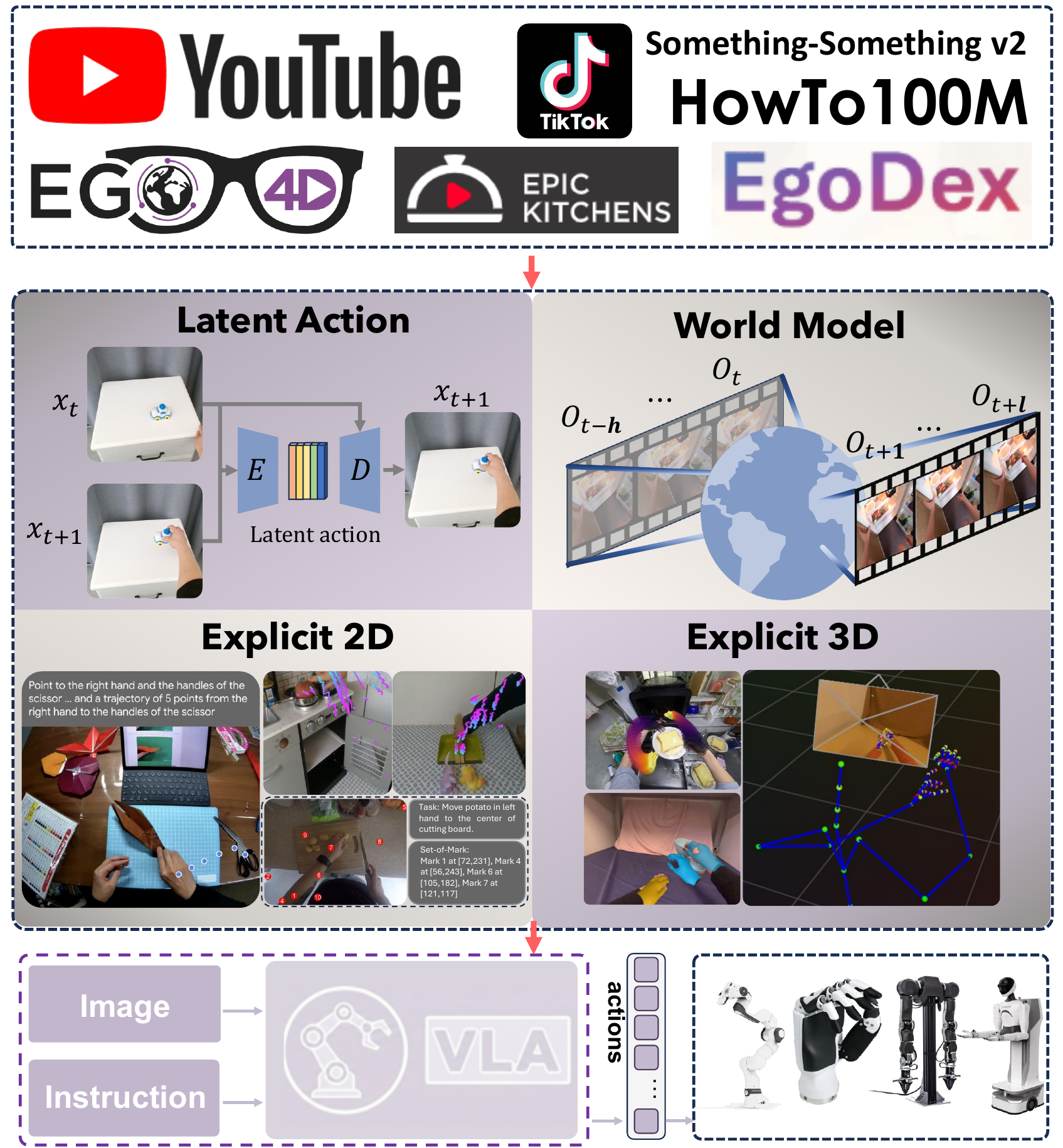}
    \caption{\textbf{Overview of scalable representation bridges for VLA models.} To leverage internet-scale human video data (top), existing methods bridge the embodiment gap via four routes: \textbf{Latent Action Abstraction}, \textbf{Predictive World Modeling}, \textbf{Explicit 2D Cues}, and \textbf{Explicit 3D Structure}. These representations transform diverse human videos into action-relevant learning signals, enabling VLA models to generate executable robot actions from image observations and language instructions (bottom).}
    \label{fig:teaser}
\end{figure}

Embodied robotic tasks require agents to perceive the world, interpret instructions, and produce temporally coherent actions under physical constraints. Recent progress in VLA models suggests that large-scale multimodal pretraining can yield increasingly generalizable robotic policies, echoing the ``foundation model'' paradigm in language and vision \cite{rt2,openvla,pi0,gr00t}. However, scaling VLA models is bottlenecked by the availability of grounded robot control data: robot demonstrations provide grounded control signals but are costly, safety-constrained, and strongly tied to a specific embodiment, sensor suite, and control frequency. Even with recent efforts to scale robot data \cite{oxe,bridgev2,droid,agibot}, robot-only datasets still under-represent long-horizon task structure, rare failure modes, and the distribution shifts in objects and environments encountered at deployment, which limits the robustness and generalization of VLA policies.

To address this bottleneck, we turn to \emph{human-centric data} as a scalable complement to robot demonstrations. 
In this survey, we are particularly interested in the visually instantiated subset of human-centric data, namely \emph{human videos}.
Compared to robot demonstrations, human videos are available at vastly larger scale---ranging from instructional videos sourced from platforms like YouTube and curated into datasets such as HowTo100M~\cite{howto100m} to diverse activities captured via lightweight wearable devices~\cite{ego4d,egoexo4d}. Beyond quantity, these resources provide rich semantic and physical cues relevant to embodied learning, such as egocentric object manipulations \cite{hoi4d,epickitchen}, multi-view procedural assemblies \cite{assembly101}, and controlled interactions with commonsense physics \cite{ssv2}. However, human videos are not “action data” for robots, they lack aligned, robot-executable action labels and proprioceptive context, and human motion does not directly map to robot kinematics and control interface. This challenge motivates the central inquiry of this survey: \emph{when human videos enter a VLA training pipeline, what type of information do they become, and how does that information interface with policy learning?}

To bridge the gap between passive human observation and active robot control, the core challenge lies in translating the rich interaction knowledge embedded in human videos into action-relevant representations compatible with robotic policy learning. In this survey, we systematize this rapidly evolving field by categorizing methods based on the \textbf{nature of the constructed representations} along the processing pipeline. From this perspective, as illustrated in Figure~\ref{fig:teaser}, we identify four distinct routes: (i) \textbf{Latent Action Abstraction}, which circumvents explicit alignment by encoding inter-frame changes or intents into discrete codes or continuous embeddings \cite{lapa,igor}; (ii) \textbf{Predictive World Modeling}, which uses video as a source of predictive learning signal to forecast future outcomes and distill these representations into policies \cite{gr1,gen2act}; (iii) \textbf{Explicit 2D Cues}, which extract interpretable image-plane signals---such as point tracks, masks, or flow---using off-the-shelf vision tools \cite{atm,magma,argos25}; and (iv) \textbf{Explicit 3D Structure}, which recovers 3D structure (poses/trajectories) to retarget motion into robot-compatible action spaces \cite{egovla,beingh0}. This taxonomy highlights the diverse \emph{representation bridges} researchers build to convert passive video observations into active control signals.

Building on this taxonomy, our survey makes three contributions:
\begin{enumerate}
    \item \textbf{Signal-centric taxonomy.} We introduce a pipeline-based taxonomy of \emph{representation bridges} that transform human videos into action-relevant representations for VLA learning, and compare the four routes---latent actions, predictive world models, explicit 2D cues, and explicit 3D structure---in terms of their representation form, scalability, and grounding requirements.
    \item \textbf{Dataset map and available signals.} We systematize representative human-video datasets along two axes---the availability of explicit geometric/3D signals and scripted vs.\ in-the-wild collection---and summarize how different supervision affordances (RGB-only vs.\ 2D/3D signals) support different transfer pipelines.
    \item \textbf{Challenges at three interfaces.} We distill open problems at three critical interfaces in the human-video-to-robot pipeline: scalable episodization of unstructured videos, heterogeneity-aware grounding under embodiment/viewpoint mismatch, and deployment-predictive evaluation of transfer efficiency.
\end{enumerate}

%% file: sections/02_background.tex
\section{Background}
\label{sec:background}
VLA learning addresses the challenge of mapping multimodal observations to executable actions for embodied agents.
At each time step $t$, the agent receives a current visual observation $o_t$, a proprioceptive state $s_t$ capturing low-dimensional robot state, such as joint configurations, gripper status, or end-effector pose, and a language instruction $l$. Modern VLA policies typically adopt a sequence-modeling formulation that conditions on the history of observations and states and predicts a chunk of future actions $a_{t:t+H}$ over a horizon $H$, promoting temporal consistency.
This process is commonly formulated as maximum-likelihood imitation learning over a dataset of robot trajectories $\mathcal{D}_{\text{robot}}$, where each trajectory $\tau$ contains aligned sequences ${(o_t,s_t,l,a_t)}_{t=1}^T$:
\begin{align}
\max_{\theta} \sum_{\tau \in \mathcal{D}_{\text{robot}}} \sum_{t} 
    \log \pi_\theta\big(a_{t:t+H} \mid o_{\le t}, s_{\le t}, l\big). 
\label{eq:imitation}
\end{align}
In practice, the action $a_t$ is instantiated as physically grounded control targets, such as end-effector motion in $\mathrm{SE}(3)$ (6-DoF pose or twist), gripper open/close (binary or continuous), arm joint commands, and, for dexterous hands, high-dimensional hand pose parameters (finger joint angles) that specify 3D interaction-relevant motion.
Standard approaches derive such action supervision from large-scale robot datasets such as Open X-Embodiment~\cite{oxe}, DROID~\cite{droid}, and AgiBot World~\cite{agibot}. However, collecting robot demonstrations at scale is prohibitively expensive and often constrained by specific hardware setups, creating a bottleneck for generalizable policy learning.

To mitigate this data scarcity, the community has turned to \emph{human-centric data} as a scalable source of interaction data.
In this survey, we consider human-centric data through the lens of VLA learning: given the formulation in Eq.~\eqref{eq:imitation}, where policies are conditioned on visual observations $o_t$, we focus on the portion of human-centric data that is available as \emph{human videos} and, within this visual setting, primarily refer to this subset simply as human videos.
Before large-scale end-to-end VLAs became prevalent, early works leveraged human videos primarily as auxiliary signals or in specialized domains. 
For example, large-scale human video datasets were used to learn robust visual representations such as R3M and MVP~\cite{r3m,mvp}, define reward functions for reinforcement learning~\cite{vip}, or extract interaction affordances to guide planning~\cite{vrb,vapo,hulc2}.

\begin{figure*}[t]
    \centering
    \includegraphics[width=0.95\textwidth]{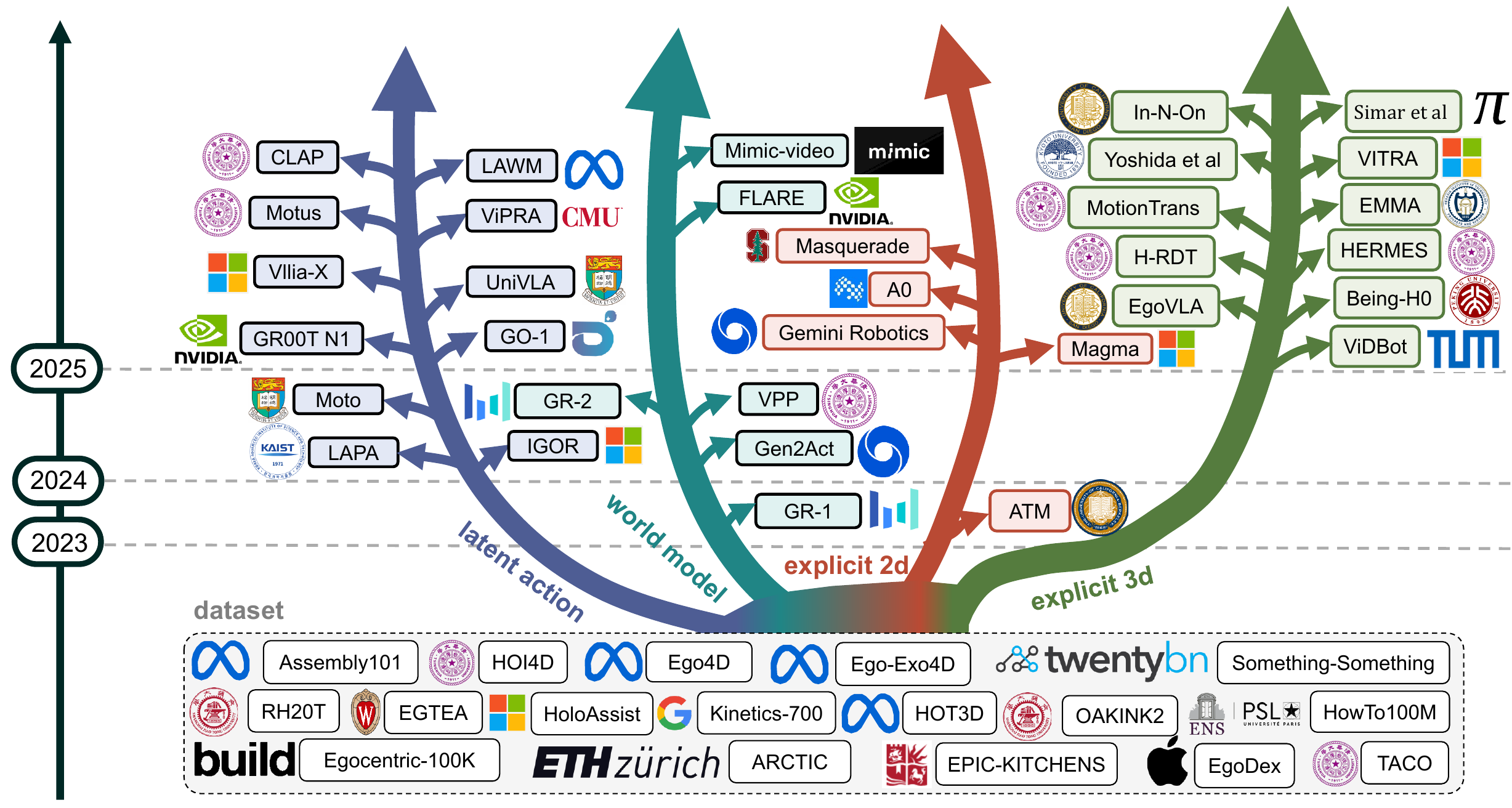} 
    \caption{\textbf{Taxonomy of Human-to-Robot Policy Transfer.}
    Representative methods are organized into four paradigms by intermediate representations:
    \textbf{Latent Actions}, \textbf{World Models}, \textbf{Explicit 2D Representations}, and \textbf{Explicit 3D Representations}.
    The bottom band illustrates commonly used large-scale human-video datasets that serve as shared foundations for these methods.}
    \label{fig:taxonomy_tree}
\end{figure*}

More recently, HERMES~\cite{hermes} formulates a unified reinforcement learning framework that transforms multi-source human motions—including raw videos—into physically plausible behaviors for mobile dexterous robots. Similarly, EMMA~\cite{emma} scales mobile manipulation by co-training on egocentric human data to reduce reliance on expensive robot teleoperation. These approaches leverage human videos to augment RL-based or modular control systems, rather than to learn generic representation bridges inside a single end-to-end VLA model.

In contrast, this survey focuses on the emerging paradigm of \textit{VLA learning from human videos}, which emphasizes the end-to-end integration of vision, language, and action within a unified generative or predictive framework.

%% file: sections/03_taxonomy.tex
\begin{table*}[t!]
\centering
\caption{Taxonomy of representation bridges from human videos to robot manipulation. We categorize existing methods based on the intermediate representation used to bridge the domain gap between human videos and robotic control. The table summarizes video sources, data scale, representations, and the end-effectors evaluated in each work. 
\textbf{Source \& Scale:} ``Source'' and ``Clips / Hours'' denote only human-video data; ``--'' indicates not reported or not applicable, and ``*'' indicates frame-based statistics.}
\label{tab:taxonomy_clean}
\small
\begin{tabularx}{\textwidth}{
    l                           
    >{\raggedright\arraybackslash\hsize=1.12\hsize}X  
    r@{ / }l                    
    >{\raggedright\arraybackslash\hsize=1.18\hsize}X 
    >{\raggedright\arraybackslash\hsize=0.7\hsize}X  
}
\toprule
\textbf{Paper} & \textbf{Source} & \multicolumn{2}{c}{\textbf{Clips / Hours}} & \textbf{Representation} & \textbf{End Effector} \\
\midrule
\rowcolor{sheader}
\multicolumn{6}{l}{\textbf{\textit{Latent Actions}}} \\

\addlinespace[0.3em]
LAPA [\citeyear{lapa}] & SSv2 & 220k & -- & VQ Latent Actions & Gripper \\
IGOR [\citeyear{igor}] & SSv2, Epic, Ego4D, EGTEA & 2m & -- & VQ Latent Actions & Gripper \\
Moto [\citeyear{moto}] & SSv2 & 105k & -- & VQ Latent Actions & Gripper \\
GO-1 [\citeyear{agibot}] & Ego4D & -- & -- & VQ Latent Actions & Gripper, Dex. Hand \\
GR00T [\citeyear{gr00t}] & Ego4D, Ego-Exo4D, Assembly-101, Epic, HOI4D, HoloAssist, RH20T-Human & -- & 2.5k & VQ Latent Actions & Gripper, Dex. Hand \\
UniVLA [\citeyear{univla}] & Ego4D & -- & -- & VQ Latent Actions & Gripper \\
Villa-X [\citeyear{villax}] & Ego4D, EgoPAT3D, EGTEA Gaze+, Epic, HO-Cap, HOI4D, HoloAssist, RH20T, SSv2 & 3.6m & -- & VQ Latent Actions & Gripper, Dex. Hand \\
ViPRA [\citeyear{vipra}] & SSv2 & 198k & -- & VQ Latent Actions & Gripper \\
Motus [\citeyear{motus}] & EgoDex & 230k & -- & Optical-Flow Latent Actions & Gripper \\
CLAP [\citeyear{clap}] & Ego4D & -- & 90h & VQ Latent Actions & Gripper \\
LAWM [\citeyear{lawm}] & YouTube Temporal-1B & -- & 13k & Sparsity/Noise-Constrained Latent Actions & Gripper \\

\midrule %
\addlinespace
\rowcolor{sheader} 
\multicolumn{6}{l}{\textbf{\textit{World Models}}} \\
\addlinespace[0.3em]
GR-1 [\citeyear{gr1}] & Ego4D & 800k & 667 & Generative Video Priors & Gripper \\
GR-2 [\citeyear{gr2}] & Ego4D, SSv2, Epic, HowTo100M, Kinetics-700 & 38m & -- & Generative Video Priors & Gripper \\
VPP [\citeyear{vpp}] & SSv2 & 192k & -- & Generative Video Priors  & Gripper, Dex. Hand \\
FLARE [\citeyear{flare}] & Self-collected & -- & -- & Future Visual Features & Gripper, Dex. Hand \\
mimic-video [\citeyear{mimicvideo}] & Web-scale & -- & -- & Generative Video Priors & Gripper, Dex. Hand \\
Gen2Act [\citeyear{gen2act}] & Web-scale & 270m+ & -- & Predicted Frames & Gripper \\
\midrule %
\addlinespace
\rowcolor{sheader} 
\multicolumn{6}{l}{\textbf{\textit{Explicit 2D}}} \\
\addlinespace[0.3em]
ATM [\citeyear{atm}] & Self-collected & 0.3k & -- & 2D Point Trajectories & Gripper \\
Magma [\citeyear{magma}] & SSv2, Epic, Ego4D & 4m & -- & 2D Point Trajectories  & Gripper \\
Gemini Robotics [\citeyear{geminirobot}] & Web-scale & -- & -- & Boxes, Keypoints, Trajectories & Gripper \\
A0 [\citeyear{A0}] & HOI4D & 22k & -- & Contact Points/Trajectories & Gripper \\
Masquerade [\citeyear{masquerade}] & Epic & 675k\textsuperscript{*} & -- & 2D Pose Trajectories & Gripper \\

\midrule %
\addlinespace
\rowcolor{sheader} 
\multicolumn{6}{l}{\textbf{\textit{Explicit 3D}}} \\
\addlinespace[0.3em]
VidBot [\citeyear{vidbot}] & Epic & -- & -- & 3D Affordance & Gripper \\
EgoVLA [\citeyear{egovla}] & HoloAssist, TACO, HOI4D, HOT3D & 500k\textsuperscript{*} & -- & 6-DoF Wrist Pose/MANO Params & Dex. Hand \\
Being-H0 [\citeyear{beingh0}] & UniHand & 2.5m & 1155 & MANO Params & Dex. Hand \\
H-RDT [\citeyear{hrdt}] & EgoDex & 338k & 829h & 6-DoF Wrist Pose/3D Fingertips  & Gripper \\
MotionTrans [\citeyear{motiontrans}] & Self-collected & 1.7k & -- & 6-DoF Wrist Pose/Hand Joints & Dex. Hand\\
Yoshida et al. [\citeyear{developing}] & Ego4D, EgoExo4D, Epic, Nymeria & 45k & -- & 6-DoF Object Pose & Gripper \\
VITRA [\citeyear{vitra}] & Ego4D, Epic, EgoExo4D, SSv2 & 1m & -- & 3D Hand Pose & Dex. Hand \\
In-N-On [\citeyear{innon}] & PH$^2$SD & -- & 1k+ & 6-DoF Wrist Pose/3D Fingertips & Dex. Hand \\
Simar et al. [\citeyear{pi05ego}] & Self-collected & -- & 14 & 3D Hand Pose & Gripper \\
\bottomrule
\end{tabularx}
\end{table*}

\section{From Video to Action: A Taxonomy of Representation Bridges}
Effectively learning VLA models from human videos necessitates a mechanism to bridge the morphology and dynamics gap between humans and robots. This is typically achieved by transforming raw video observations into structured \emph{intermediate representation bridges}—ranging from explicit 3D hand poses to abstract latent vectors—which distill control-relevant semantics while discarding action-irrelevant visual variation.

To provide a structured overview of this evolving landscape, we categorize representative methods into four paradigms based on their choice of intermediate representation (Fig.~\ref{fig:taxonomy_tree}). We further summarize these works in Table~\ref{tab:taxonomy_clean}, detailing their human video data sources, data scale, and the types of robotic end-effectors used for evaluation. This taxonomy highlights a rapidly diversifying space: explicit geometric representations continue to drive progress in high-precision dexterous control, and in parallel, implicit latent representations and predictive world models are increasingly explored to harness the vast potential of internet-scale, unlabelled human videos.

\subsection{Latent Actions}
Latent-action learning methods aim to derive compact action representations from videos in a self-supervised manner, thereby circumventing the high cost of acquiring additional action annotations. A common paradigm is to take observations at two time steps, $o_t$ and $o_{t+H}$, as input and infer a latent vector $z_t$, such that the future observation $o_{t+H}$ can be reconstructed from $z_t$ and $o_t$. 
By introducing a strong information bottleneck (\emph{e.g.}, a VQ-VAE–style discretization~\cite{van2017neural}), $z_t$ is forced to capture only the most salient inter-frame changes associated with action-relevant dynamics, effectively serving as a proxy for the action variable in Eq.~\eqref{eq:imitation} when absorbing action knowledge from human videos.
Since its introduction in VLA frameworks such as \textbf{LAPA}~\cite{lapa}, key challenges have been: (i) learning a compact, expressive representation of action-relevant dynamics that generalizes across diverse embodiments and data sources; and (ii) effectively integrating latent action prediction into the VLA training pipeline. 

For learning latent actions, \textbf{IGOR}~\cite{igor} applies asymmetric cropping augmentations during encoding and decoding to discourage incorporating absolute positional information. \textbf{UniVLA}~\cite{univla} replaces pixel-space supervision with DINOv2~\cite{dinov2} features to avoid learning noisy details, and introduces a two-stage discretization with the aid of task instruction to disentangle task-irrelevant motions (\emph{e.g.}, camera motions). \textbf{ViPRA}~\cite{vipra} enforces optical flow consistency loss during reconstruction to encourage physically plausible motion encoding, whereas \textbf{Motus}~\cite{motus} takes optical flow between observations as direct input, explicitly reducing the influence of appearance variations in latent action encoding. Some methods further incorporate robot data with ground-truth actions to better align latent actions with robot control. For example, \textbf{villa-X}~\cite{villax} jointly learns human and robot latent action spaces conditioned on embodiment context, leveraging labeled robot data to ground the latent space in action-conditioned dynamics. Similarly, \textbf{CLAP}~\cite{clap} aligns human-video latents with a quantized robot-derived latent space via contrastive learning, mitigating entanglement with irrelevant visual changes. Moving beyond discrete tokens, \textbf{LAWM}~\cite{lawm} adopts continuous actions with sparsity constraints to better capture large-scale, in-the-wild action dynamics.

From another perspective, several works explore how to integrate latent action prediction into VLA training. \textbf{Moto}~\cite{moto} adopts an autoregressive pretraining scheme for next latent action prediction, followed by inserting additional query tokens for robot action decoding during finetuning. \textbf{GO-1}~\cite{agibot} employs a hierarchical design, learning a latent planner that predicts future latent actions to bridge the vision-language backbone and a flow-based action expert. \textbf{GR00T N1}~\cite{gr00t} uses latent actions as pseudo-targets for action decoding on human demonstrations, while jointly training with real action labels from heterogeneous robot data. 

Nevertheless, despite their effectiveness in extracting knowledge from human videos, it remains an open question whether highly compressed latent tokens can adequately represent the rich structure of complex, high-DoF dexterous manipulation.

\subsection{World Models}
\label{sec:world_models}
World-model-based approaches aim to predict how the environment evolves over time. For human videos, this encompasses not only the future actions of the human agent but also changes induced by its interactions with the environment. The underlying assumption is that accurately forecasting such dynamics equips the model with the action-relevant information needed for downstream control. Formally, this process can be expressed as $p(S_{t+1:t+H} \mid o_{\le t}, l)$, where $S_{t+1:t+H}$ denotes future environment state (\emph{e.g.}, raw pixels or visual features), conditioned on past observations $o_{\le t}$ and the language instruction $l$\footnote{A stricter variant further conditions on future actions; in this survey, we focus on approaches that do not rely on action inputs.}. This formulation can be learned via video generation without action supervision, analogous to latent actions but potentially more expressive for complex dynamics. 

Existing methods mainly study world-model architectures and how to transfer their learned knowledge to VLA action prediction.
An early representative is \textbf{GR-1}~\cite{gr1}, which proposes a GPT-style world model that autoregressively predicts future frames conditioned on the language instructions and visual observations. The model is pretrained on egocentric human videos, and subsequently fine-tuned on robot datasets with an additional next-action prediction objective, leveraging internal features learned during frame generation.
This paradigm is scaled by \textbf{GR-2}~\cite{gr2} using web-scale human data~\cite{howto100m} and VQGAN-based~\cite{vqgan} visual tokens to improve the fidelity of future frame prediction, thereby facilitating the acquisition of more accurate action-related knowledge. \textbf{FLARE}~\cite{flare} jointly trains its action decoder to perform action prediction and implicit world modeling by enforcing representation alignment (REPA)~\cite{yu2024representation} between decoder's internal features and future visual features on action-unlabeled human video.
In contrast, VPP~\cite{vpp} treats a pretrained video generation model (SVD~\cite{svd}) as a frozen feature extractor, using its ``predictive visual representations'' as conditions for downstream action policies.
More recently, \textbf{Mimic-Video}~\cite{mimicvideo} also shows that action prediction can be enhanced by leveraging intermediate representations from partially denoised steps of a pretrained video diffusion model as conditioning signals.

A complementary direction uses video generation as an explicit planning interface. 
\textbf{Gen2Act}~\cite{gen2act} employs a pretrained video generation model to synthesize a ``hallucinated'' video plan from a static image. It constructs paired data by conditioning the generator on the initial frame of each robot demonstration to produce a corresponding human-like video, and then trains a video-conditioned policy that maps generated plans to robot actions.

Despite their promising results, world-model-based methods often incur substantial training overhead, since learning world models requires generating large numbers of visual tokens. In addition, effectively distilling and transferring the knowledge embedded in world model to action prediction remains a non-trivial and under-explored challenge.

\subsection{Explicit 2D Representations}
\label{sec:explicit2d}
A complementary line of work relies on 2D visual primitives (\emph{e.g.}, keypoints, bounding boxes, 2D point trajectories) as intermediate representations. 
Unlike self-supervised latent actions or world models, these approaches leverage explicit and interpretable spatiotemporal cues, often extracted using 2D vision models, to bridge perception and action learning. The extracted cues are typically used as additional supervision to facilitate the acquisition of action-relevant knowledge from human videos.

For example, \textbf{ATM}~\cite{atm} uses dense point trajectories as action-like supervision, training a transformer to predict future motion, which then serves as a dynamic spatial prior for downstream manipulation.
\textbf{Magma}~\cite{magma} turns trajectories into an in-context control cue by rendering them as ``Trace-of-Mark'' visual prompts, effectively converting unlabeled videos into VLA-style training pairs with explicit temporal structure.
At larger scale, \textbf{Gemini Robotics}~\cite{geminirobot} trains an embodied reasoning backbone on human videos annotated with diverse 2D primitives (boxes, keypoints, and tracks), leveraging these signals to improve long-horizon generalization.

Beyond generic motion cues, \textbf{A0}~\cite{A0} focuses on explicit spatial-affordance supervision by modeling object-centric contact points and post-contact trajectories as embodiment-agnostic interaction representations, enabling robust spatial reasoning across heterogeneous embodiments. From a complementary perspective, \textbf{Masquerade}~\cite{masquerade} reduces the human–robot embodiment gap by editing in-the-wild human videos to visually resemble robot demonstrations, providing explicit 2D robot keypoint supervision that improves data efficiency and long-horizon generalization.

While these methods efficiently exploit existing vision tools to extract action-relevant cues, they may inherit label noise from the underlying models and often require additional filtering. Furthermore, 2D cues can be fundamentally limited in expressing complex manipulation, due to the lack of depth information and occlusion reasoning, as real-world interactions are inherently three-dimensional.

\begin{table*}[ht]
\centering
\caption{\textbf{Classification of Human-Video Datasets for VLA.} \textbf{Scale} reports dataset size using the most standard unit for each dataset. \textbf{Geo Signal} indicates available geometric/3D-related signals (not necessarily explicit 3D labels). \textbf{Scripted} denotes whether data collection follows predefined tasks/procedures (staged/structured) rather than fully in-the-wild recording.}
\label{tab:vla_dataset_revised}
\small
\setlength{\tabcolsep}{8pt}
\renewcommand{\arraystretch}{1.15}
\begin{tabular}{@{} l r c c c c @{}}
\toprule
\textbf{Dataset} & \textbf{Scale} & \textbf{View} & \textbf{Language Annotation} & \textbf{Geo Signal} & \textbf{Scripted} \\
\midrule

\rowcolor[gray]{0.95} \multicolumn{6}{l}{\textit{Category 1:  Datasets w/o Explicit 3D Action Labels}} \\
Something-Something V2 (SSv2)~\cite{ssv2} & 220K clips & 3rd & \cmark & RGB & \cmark \\
EPIC-KITCHENS-100 (Epic)~\cite{epickitchen} & 100 hours & Ego & \cmark & RGB & \xmark \\
Ego4D~\cite{ego4d} & 3.6K hours & Ego & \cmark & RGB & \xmark \\
EgoExo4D~\cite{egoexo4d} & 1.3K hours & MV + Ego & \cmark & MV (calib) & \xmark \\
HowTo100M~\cite{howto100m} & 136M clips & Mix & \cmark & RGB & \xmark \\
Egocentric-100K~\cite{buildaiegocentric100k2025} &  100K hours & Ego & \xmark & RGB & \xmark \\
EGTEA Gaze+~\cite{egtea} & 28 hours & Ego & \cmark & Gaze + Hand & \cmark \\
RH20T~\cite{rh20t} & 110K clips & MV + Ego & \cmark & MV (calib) & \cmark \\
Kinetics-700~\cite{kinetics700} & 650K clips & 3rd & \cmark & RGB & \xmark \\
\midrule

\rowcolor[gray]{0.95} \multicolumn{6}{l}{\textit{Category 2: Datasets w/ Explicit 3D Action Labels}} \\
HOI4D~\cite{hoi4d} & 2.4M frames & Ego & \xmark & RGB-D & \cmark \\
Assembly-101~\cite{assembly101} & 513K clips & MV + Ego & \cmark & MV (calib) & \cmark \\
EgoDex~\cite{hoque2025egodex} & 300K episodes & Ego & \cmark & RGB & \cmark \\
OAKINK2~\cite{oakink2} & 4.0M frames & MV + Ego & \cmark & MV (calib) & \cmark \\
ARCTIC~\cite{fan2023arctic} & 2.1M frames& MV + Ego & \cmark & MV (calib) & \cmark \\
TACO~\cite{taco}& 5.2M frames & MV + Ego & \xmark & MV (calib) & \cmark \\
HoloAssist~\cite{holoassist} & 166 hours & MV + Ego & \cmark & RGB-D & \cmark \\
HOT3D~\cite{hot3d} & 3.7M images & Ego & \xmark & MV (calib) & \cmark \\
\bottomrule
\end{tabular}
\end{table*}

\subsection{Explicit 3D Representations}
\label{sec:explicit3d}
Similar in spirit to 2D-based approaches, recent methods leverage 3D information extracted from human videos to train VLA policies.
Compared to 2D cues, 3D representations provide a coordinate-consistent interface  (\emph{e.g.}, SE(3) trajectories, parametric hand states, object poses), reducing viewpoint dependence and depth ambiguity, and aligning more naturally with VLA action prediction.
In many systems, \textbf{parametric hand models} (\emph{e.g.}, MANO~\cite{mano}) serve as a canonical and compact 3D representation of human manipulation and are widely used as a shared kinematic space linking human videos to robot control. Methods primarily vary in the types of videos they exploit, the strategies used to estimate 3D labels, and the mechanisms for bridging human and robot action spaces.

\textbf{EgoVLA}~\cite{egovla} leverages videos with high-quality 3D hand action annotations, obtained via multi-view optimization or RGB-D SLAM, to pretrain VLA models. It unifies human and robot action prediction by considering only wrist poses and fingertip positions, and adopts inverse kinematics from this human action space for robot execution. \textbf{H-RDT}~\cite{hrdt} adopts a similar hand-centric representation and leverages videos with hand annotations collected using AR/VR devices. \textbf{In-N-On}~\cite{innon} also leverages AR-captured human videos and incorporates additional head poses into the unified action representation.
\textbf{Being-H0}~\cite{beingh0} scales to larger human-video corpora and discretizes continuous MANO hand trajectories into a vocabulary of motion tokens for VLA pretraining. 

Beyond learning from videos captured in controlled settings, recent methods have explored recovering 3D information from more in-the-wild videos. 
\textbf{VidBot}~\cite{vidbot} combines structure-from-motion with depth foundation models to recover 3D affordance trajectories from unscripted videos. 
\textbf{VITRA}~\cite{vitra} proposes an automated pipeline that converts large collections of unscripted real-world human videos into robot-aligned atomic action segments with MANO hand annotations and language instructions, enabling more scalable pretraining. 
\textbf{Yoshida et al.}~\cite{developing} instead reconstruct 3D object pose changes from egocentric videos as supervisions.

In addition to the human-pretraining and robot-fine-tuning paradigm, recent work has investigated alternative learning schemes for VLA.
For example, \textbf{MotionTrans}~\cite{motiontrans} and work by \textbf{Kareer et al.}~\cite{pi05ego} collect human demonstrations with 3D action annotations using wearable devices and jointly train VLA models on both human and robot data, showing that such co-training improves robots' generalization to previously unseen tasks.

Although 3D representations are better aligned with VLA action prediction and facilitate human-to-robot transfer, acquiring accurate 3D labels remains difficult.
Wearable devices can provide high-quality annotations in controlled settings, but web-scale data must rely on 3D reconstruction from vision models, which is particularly challenging in in-the-wild scenarios and often introduces larger errors than 2D supervision. Moreover, directly predicting actions in 3D space requires models to not only understand image-space observations but also infer the underlying 3D world structure from 2D projections.
From a generalization perspective, whether this paradigm is ultimately more effective than predicting in 2D image space remains an open question.

%% file: sections/04_datasets.tex
\section{Human-Centric Data for VLA Learning}
In this section, we provide a detailed review of the human-centric data sources underpinning the aforementioned approaches, with a focus on their visual instantiation in the form of human videos. 
We organize them along two axes: whether they provide \emph{explicit metric 3D action labels} and whether collection is \emph{scripted} or \emph{unscripted}. Explicit 3D action labels determine whether action supervision can be directly grounded into robot-executable actions or are typically inferred from visual state transitions~\cite{blank2024scaling,hirose24lelan}. Table~\ref{tab:vla_dataset_revised} summarizes this taxonomy.

\paragraph{Datasets without Explicit 3D Action Labels.}
These datasets provide RGB videos with semantic annotations (e.g., scenes, objects, action descriptions) but lack metric 3D trajectories of hands or bodies in a camera/world frame, such as 3D keypoints or MANO parameters. Consequently, they are mainly used to support \emph{latent-action} or \emph{world-model}-based approaches, where actions are inferred from visual state transitions rather than directly specified in a 3D action space.

Representative datasets include Something-Something V2 (short hand–object clips for manipulation dynamics) \cite{ssv2}, EPIC-KITCHENS (large-scale egocentric kitchen activities) \cite{epickitchen}, Ego4D (egocentric daily-life video across diverse scenarios) \cite{ego4d}, HowTo100M (Internet-scale instructional video) \cite{howto100m}, Egocentric-100K (100K hours from factory environments, largely without annotations) \cite{buildaiegocentric100k2025}, and Ego-Exo4D (synchronized ego–exo multi-view recordings for cross-view learning) \cite{egoexo4d}. Despite their scale and diversity, these datasets are typically used as video sources rather than as providers of metric 3D action representations.
In terms of collection protocol, Something-Something V2 is scripted via predefined action templates \cite{ssv2}, whereas EPIC-KITCHENS, Ego4D, Ego-Exo4D, HowTo100M, and Egocentric-100K are unscripted \cite{epickitchen,ego4d,egoexo4d,howto100m,buildaiegocentric100k2025}. The absence of predefined boundaries in continuous in-the-wild recordings complicates both action segmentation and instruction annotation for imitation learning.

\paragraph{Datasets with Explicit 3D Action Labels.}
A smaller but crucial class of datasets provides \emph{explicit metric 3D action annotations}, including 3D keypoint trajectories of hands/bodies and parametric representations such as MANO or SMPL. They are typically captured with instrumented setups (e.g., AR/VR headsets, depth sensors, or motion capture) that record motion in camera/world coordinates. 

Representative datasets in this domain include HOI4D \cite{hoi4d}, HOT3D \cite{hot3d}, EgoDex \cite{hoque2025egodex}, ARCTIC \cite{fan2023arctic}, TACO \cite{taco}, Assembly101 \cite{assembly101}, OakInk2 \cite{oakink2} and HoloAssist \cite{holoassist}. These works advance egocentric perception across varying levels of granularity and task complexity. In terms of 3D perception, HOI4D provides egocentric RGB-D with frame-wise 3D hand and object poses \cite{hoi4d}, while HOT3D further addresses occlusion challenges by leveraging egocentric multi-view videos to achieve high-fidelity hand and object tracking \cite{hot3d}. Regarding fine-grained manipulation, EgoDex uses head-mounted AR to collect 3D hand trajectories for dexterous imitation learning \cite{hoque2025egodex}, ARCTIC captures bimanual articulation with dynamic contact information \cite{fan2023arctic}, and Taco establishes a benchmark for generalizable bimanual tool-action-object understanding \cite{taco}. At the task and assistance level, Assembly101 focuses on long-horizon procedural activities with dense segmentation \cite{assembly101}, OakInk2 adopts an object-centric hierarchy (affordances to complex tasks) for structured motion generation \cite{oakink2}, and HoloAssist targets interactive AI assistants with real-world data centered on human intent and guidance \cite{holoassist}.
These datasets enable direct supervision with explicit 2D/3D action labels and are closest to robot imitation learning, often improving data efficiency \cite{vitra}. However, they are usually smaller-scale and collected in laboratory or instrumented settings, trading diversity for annotation fidelity. Consistent with this, collection is commonly \emph{scripted}, with predefined interactions and objects and carefully segmented action units that align well with imitation-learning supervision, but limit behavioral diversity, scalability, and representativeness relative to large unscripted corpora.

%% file: sections/05_challenges.tex
\section{Challenges and Future Directions}
\label{sec:challenges}
In this section, we analyze open challenges shared across methods that leverage human videos, organizing them around three key interfaces in the human-video-to-robot pipeline: transforming unstructured videos into training-ready episodes, handling heterogeneity when mapping video-derived supervision to robot-executable actions, and designing evaluations that predict real deployment performance.

\subsection{Transforming Human Videos to Episodes}
\label{subsec:video2episode}
A bottleneck appears upstream of policy learning. Web-scale human videos from platforms such as YouTube and TikTok are rarely packaged into episodes that match the supervision granularity required by robot learning. Instructional corpora such as HowTo100M \cite{howto100m} offer a scalable proxy but suffer from weak alignment between narration and manipulation, while egocentric datasets \cite{ego4d,egoexo4d,epickitchen} remain largely untrimmed and often contain viewpoint changes and off-task segments. As a result, clips frequently mix intents, and similar visual transitions can correspond to different latent actions. This ambiguity degrades learning signals by destabilizing discrete tokenization and diverting model capacity toward control-irrelevant dynamics. Procedural datasets such as HOI4D and Assembly101 \cite{hoi4d,assembly101} mitigate this issue by aligning video units with manipulation primitives, whereas in-the-wild video requires explicit episodization, as explored by systems such as VITRA \cite{vitra}. Future episodization would benefit from semantic- and interaction-driven segmentation, defining segments around manipulation phases and object state changes rather than fixed time windows. Standardized construction protocols are also needed to reduce hidden variance and support rigorous comparison.

\subsection{Addressing Heterogeneity}
\paragraph{Embodiment Mismatch.}
\label{subsec:embodiment_mismatch}
Embodiment mismatch stems from kinematic and contact-mechanics differences between human hands and robotic end-effectors. Retargeting is intrinsically under-constrained when high-DoF human motion is mapped onto lower-DoF robot control spaces. As a result, distinct human motions can yield the same robot command, and some motions may admit no feasible realization~\cite{dexmv}. This issue is amplified for simple grippers and across diverse end-effector morphologies, where action spaces are not aligned, complicating cross-morphology VLA training~\cite{hrdt}. A practical response is to emphasize embodiment-invariant supervision, such as interaction intent and object-centric effects, rather than literal trajectory imitation. Hardware advances may reduce the gap~\cite{sharpawave,optimus}, but reliable transfer will still require morphology-conditioned grounding mechanisms that map shared priors into feasible commands under platform-specific constraints.


\paragraph{Viewpoint Heterogeneity.}
\label{subsec:viewpoint_heterogeneity}
Robot manipulation policies are typically trained and deployed with fixed, robot-centric views, such as head-mounted cameras, often supplemented by wrist or eye-in-hand cameras for contact-centric detail~\cite{pi0}. In contrast, web-scale human video, however, is predominantly exocentric, where fine contacts are weakly observed and occlusions around the interaction site are common \cite{howto100m}. Egocentric human video is closer but still mismatched, since head-mounted streams (\emph{e.g.}, Ego4D, EPIC-KITCHENS) exhibit rapid viewpoint shifts driven by head turns and frequent occlusions, which can remove active hand–object contact from view during manipulation \cite{ego4d,epickitchen}. Progress will likely require explicit cross-view alignment between human and robot observations and an action representation that is anchored to interaction outcomes, for example, object state changes, rather than view-specific appearance.

\subsection{Benchmarking and Evaluation}
\label{subsec:eval}
A key challenge is that existing benchmarks only partially reflect the regime in which human video supervision is intended to help. Standard suites such as LIBERO~\cite{libero}, CALVIN~\cite{calvin}, and SIMPLER~\cite{simpler} support controlled comparison, but their task templates and simulated scenes under-represent the diversity and long-horizon structure typical of internet human videos. As a result, reported improvements can be benchmark-specific and do not directly answer the central question, namely transfer efficiency under a fixed robot-data budget. Real-robot evaluation remains necessary, yet scaling and standardizing it is difficult due to cost, safety, and platform variability.
Progress will benefit from evaluation protocols that explicitly track deployment-relevant generalization and transfer efficiency. This includes reporting performance as a function of robot-data budget, using held-out splits over novel objects and scenes, and quantifying robustness under viewpoint and embodiment shifts. In particular, to isolate the contribution of human video pretraining, evaluation protocols should compare models with and without human-video-based initialization under matched robot-data and compute budgets across diverse downstream tasks and embodiments. Testbeds closer to human activity distributions, such as BEHAVIOR-1K~\cite{behavior1k}, together with scalable digital-twin and domain-randomized toolchains such as Cosmos~\cite{cosmos} and RoboTwin~\cite{robotwin2}, can increase coverage and reproducibility. Complementary efforts on LLM-driven task generation in simulation, such as GenSim and GenSim2~\cite{gensim,gensim2}, suggest a path toward automatically constructing more diverse and long-horizon evaluation suites, although these approaches currently operate largely without directly leveraging human videos.

%% file: sections/06_conclusion.tex
\section{Conclusion}
\label{sec:conclusion}
Human videos offer a scalable route to easing the data bottleneck in VLA learning, but only when the gap between human observation and robot execution is bridged in a principled way. In this survey, we organized recent progress around a taxonomy of representation bridges that transform human videos into action-relevant signals, covering approaches that infer latent actions, learn predictive world models, and construct explicit 2D and 3D representations. This taxonomy highlights complementary strengths, with implicit representations scaling across diverse video sources and explicit geometric cues offering stronger grounding and a clearer interface to robot-executable actions.

Looking ahead, progress will depend not only on better representations, but also on the surrounding pipeline, including episodizing in-the-wild video, handling kinematic and viewpoint mismatch, and developing evaluation protocols that better predict deployment performance and transfer efficiency. Addressing these challenges is central to making broadly pretrained VLA models, learned from human videos, generalize reliably across tasks, embodiments, and real-world environments.